\documentclass[runningheads]{llncs}
\usepackage[misc]{ifsym}
\usepackage{graphicx}
\usepackage{amsmath}
\usepackage{amssymb}
\usepackage{booktabs}
\usepackage{amssymb}
\usepackage{subcaption}
\usepackage[pagebackref,breaklinks,colorlinks]{hyperref}
\usepackage[capitalize]{cleveref}
\crefname{section}{Sec.}{Secs.}
\Crefname{section}{Section}{Sections}
\Crefname{table}{Table}{Tables}
\crefname{table}{Tab.}{Tabs.}

\begin{document}
\title{Overcoming the Limitations of Localization Uncertainty: 
Efficient \& Exact Non-Linear Post-Processing and Calibration}
\titlerunning{Overcoming the Limitations of Localization Uncertainty}
%
\author{Moussa {Kassem Sbeyti}\inst{1,2}(\Letter) \and
Michelle Karg\inst{1} \and
Christian Wirth\inst{1} \and \\
Azarm Nowzad\inst{1} \and
Sahin Albayrak\inst{2}}
\authorrunning{M. Kassem Sbeyti et al.}
%
\institute{Continental AG, Germany \\
\email{\{moussa.kassem.sbeyti, michelle.karg,\\christian.2.wirth, azarm.nowzad\}@continental-corporation.com} \and DAI-Labor, Technische Universit{\"a}t Berlin, Germany \\
\email{sahin.albayrak@dai-labor.de} 
}
\maketitle              
\begin{abstract}
Robustly and accurately localizing objects in real-world environments can be challenging due to noisy data, hardware limitations, and the inherent randomness of physical systems. To account for these factors, existing works estimate the aleatoric uncertainty of object detectors by modeling their localization output as a Gaussian distribution $\mathcal{N}(\mu,\,\sigma^{2})\,$, and training with loss attenuation. We identify three aspects that are unaddressed in the state of the art, but warrant further exploration: (1) the efficient and mathematically sound propagation of $\mathcal{N}(\mu,\,\sigma^{2})\,$ through non-linear post-processing, (2) the calibration of the predicted uncertainty, and (3) its interpretation. We overcome these limitations by: (1) implementing loss attenuation in EfficientDet, and proposing two deterministic methods for the exact and fast propagation of the output distribution, (2) demonstrating on the KITTI and BDD100K datasets that the predicted uncertainty is miscalibrated, and adapting two calibration methods to the localization task, and (3) investigating the correlation between aleatoric uncertainty and task-relevant error sources. Our contributions are: (1) up to five times faster propagation while increasing localization performance by up to 1\%, (2) up to fifteen times smaller expected calibration error, and (3) the predicted uncertainty is found to correlate with occlusion, object distance, detection accuracy, and image quality.
\end{abstract}

\section{Introduction}
\label{sec:introd}
Object detectors in safety-critical systems face multiple challenges, including limited sensor resolution, difficult weather conditions, and ambiguous situations \cite{kraus2019uncertainty,feng2018towards,he2019bounding}. These challenges decrease performance regardless of the training frequency, as they induce an inevitable uncertainty called aleatoric uncertainty \cite{kendall2017uncertainties}. Therefore, existing works explicitly integrated aleatoric uncertainty into object detectors via loss attenuation \cite{kendall2017uncertainties} for varying applications, such as enhancing safety, robustness, and performance \cite{choi2019gaussian,kraus2019uncertainty,le2018uncertainty}. This paper prioritizes localization due to the absence of confidence information from the localization head in object detectors, when compared to the scores provided by the classification head.

EfficientDet \cite{tan2020efficientdet}, a one-stage anchor-based detector, demonstrates state-of-the-art performance in terms of both accuracy and speed on various benchmark datasets, making it an ideal use-case for this paper. An anchor-based detector predicts anchor-relative offsets, which are subjected to non-linear transformations during post-processing to compute the final object coordinates. These offsets are modeled as distributions to account for uncertainty, which raises the crucial question: \textbf{How is the output distribution, including the uncertainty, propagated through non-linear functions?} Le et al.~\cite{le2018uncertainty} is the only work known to us, that considers the propagation of the anchor-relative offsets through non-linearities, and addresses it via sampling from the estimated distribution. However, sampling has the downside of either a high computation time for a large sample size or a reduced accuracy for a small sample size. We therefore develop two novel, fast and exact approaches. The first method is based on normalizing flows, with the main advantage of a universal applicability to many non-linear, arbitrarily complex functions and output distributions. The second method is tailored towards a normal output distribution $\mathcal{N}(\mu,,\sigma^{2}),$ transformed by an exponential function. It utilizes the properties of the log-normal distribution, and its main advantage is an efficient usage of computational resources. 

Once the uncertainty is propagated, the focus shifts to assessing its quality: \textbf{Is the predicted localization uncertainty well-calibrated?} Other research on localization uncertainty estimation in object detection typically overlooks its calibration \cite{he2019bounding,le2018uncertainty,harakeh2020bayesod,choi2019gaussian,kraus2019uncertainty}. Hence, we introduce different approaches to calibrate it, inspired by calibration for general classification and regression tasks. We select two established methods: calibrating via an auxiliary model, e.g.~isotonic regression \cite{vv,zadrozny2002transforming,kuleshov2018accurate}, and factor scaling \cite{guo2017calibration,laves2021recalibration}. We extend the first method to coordinate- and class-specific calibration. For the second calibration method, we establish and evaluate various loss functions during the optimization phase of the scaling factor, which directly adjusts the predicted uncertainty by considering its proximity to the residuals. Both methods are further improved by incorporating the object size, where each object's uncertainty is normalized by its width and height, resulting in a balanced calibration of objects of all sizes and aspect ratios. Furthermore, we provide a data selection process for calibration, which allocates all predictions to their ground-truth based on proximity, in contrast to, e.g.~thresholding detections based on the classification score.  

\begin{figure}
  \centering
   \includegraphics[width=\textwidth]{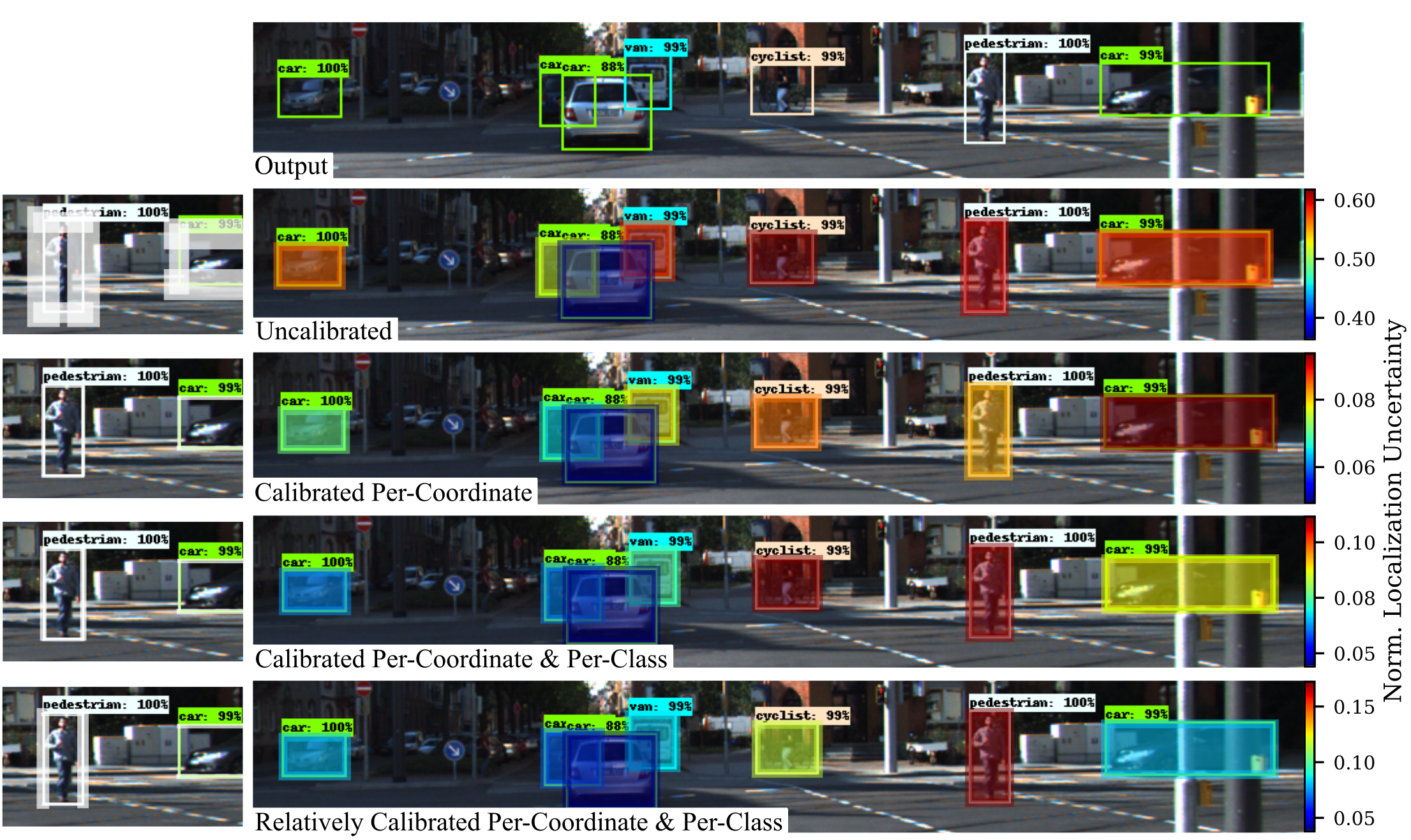}
    \caption{Aleatoric localization uncertainty $\sigma$ \textbf{per object} normalized by its width and height (right), and an example crop with $\mu\pm10\sigma$ \textbf{per coordinate} (left). Uncertainty correlates with occlusion, distance and detection performance. Calibration reduces the uncertainty. Per-coordinate calibration strengthens the correlation with localization accuracy. Per-class calibration shifts the uncertainty towards classes with lower detection accuracy. Relative calibration accounts for the impact of the different object areas and aspect ratios. }
   \label{fig:godfig}
\end{figure}

After the localization uncertainty is estimated, propagated and calibrated, its interpretability is required to define potential applications (see \cref{fig:godfig}): \textbf{What correlations exist between the data and the uncertainty?} Related works discover that aleatoric uncertainty correlates with occlusion \cite{feng2018towards,feng2019leveraging,choi2019gaussian,kraus2019uncertainty} and object distance due to sparsity in point clouds \cite{feng2018towards}, but not with detection accuracy \cite{feng2018towards,feng2019leveraging}. We investigate the latter and discover the contrary. We verify and show to which extent aleatoric uncertainty correlates with occlusion and detection performance, and extend the analysis to the object area in an image, i.e.~object distance, and the quality of the image cropped around each detection.

In summary, the contributions of our work are:
\begin{itemize}
\item Development of two novel, exact and fast methods for uncertainty propagation through non-linear functions, enabling accurate uncertainty estimation without additional drawbacks.
\item Development and extension of two calibration methods and a data selection approach for accurate calibration in the context of object localization.
\item A comprehensive experimental overview of the quality and correlation between aleatoric uncertainty and traceable metrics, which further advances the understanding of aleatoric uncertainty.
\end{itemize}
\section{Background and Related Work}
\label{sec:relw}
This section presents a concise overview of existing works on aleatoric uncertainty estimation, decoding in object detectors, and calibration for regression.

\textbf{Loss Attenuation.} A widely adopted approach for estimating aleatoric uncertainty is the sampling-free loss attenuation, which assumes that observation noise is dependent on the input \cite{kendall2017uncertainties}. By extending the network output to include both the mean $\mu$ and variance $\sigma^2$, i.e.~modeling it as a Gaussian distribution, and training the network on the negative log-likelihood (NLL), the uncertainty can be learned as a function of the data. Choi et al.~\cite{choi2019gaussian}, Kraus and Dietmayer \cite{kraus2019uncertainty} and Feng, Rosenbaum and Dietmayer \cite{feng2018towards,feng2019leveraging} show that loss attenuation enhances the performance of 2D and 3D object detectors. They find that the estimated uncertainty correlates with occlusion \cite{kraus2019uncertainty,feng2018towards,feng2019leveraging} and object distance based on LiDAR data \cite{feng2018towards,feng2019leveraging}, but it does not correlate with detection accuracy, measured via the intersection over union (IoU) \cite{feng2018towards,feng2019leveraging}. The focus of these works is primarily performance enhancement of object detectors, as they place less emphasis on the reliability and interpretability of the uncertainty estimates.

\textbf{Anchor-Relative Localization.} Choi et al.~\cite{choi2019gaussian} and Kraus and Dietmayer \cite{kraus2019uncertainty} implement loss attenuation in YOLOv3 \cite{redmon2018yolov3}. Anchor-based object detectors such as YOLOv3 \cite{redmon2018yolov3}, single-shot detector (SSD) \cite{liu2016ssd}, and EfficientDet \cite{tan2020efficientdet} divide their final feature maps into a grid. Whereby each grid cell contains a pre-defined set of static bounding boxes known as anchors. During training, the detector learns the offsets for the center, width and height between the pre-defined anchors and the ground truth. In the post-processing, the predicted offsets are decoded based on their corresponding anchors, usually via non-linear functions, such as exponential and sigmoid \cite{tan2020efficientdet,liu2016ssd,redmon2018yolov3,redmon2016you}. This transforms them into bounding box coordinates, which are then scaled to the original image size. As introduced in \cref{sec:introd}, Le et al.~\cite{le2018uncertainty} is the only work that considers the non-linearity in the decoding process. They implement loss attenuation in SSD \cite{liu2016ssd}. To decode the anchor-relative coordinates along their corresponding variances, they draw samples from the predicted multivariate normal distribution $\mathcal{N}(\mu,\,\sigma^{2})\,$, decode the samples, then calculate the mean and variance of the decoded values. Other works do not explicitly address the non-linearity in the decoding process, i.a. decode the predicted variance by reversing the encoding equation of the mean \cite{choi2019gaussian,kraus2019uncertainty,he2019bounding,harakeh2020bayesod}. Therefore, there is currently no deterministic and exact method available for decoding the values of both $\mu$ and $\sigma^2$.

\textbf{Regression Uncertainty Calibration.} Calibration is crucial after estimating and propagating the uncertainty. Approximate Bayesian approaches such as loss attenuation produce miscalibrated uncertainties \cite{feng2021review}. Laves et al.~\cite{laves2021recalibration} and Feng et al.~\cite{feng2019can} argue that minimizing the NLL should result in the estimation of $\sigma^2$ matching the squared error. However, they and Phan et al.~\cite{phan2018calibrating} find that the prediction of $\sigma^2$ is in reality biased, since it is predicted relative to the estimated mean. Kuleshov, Fenner and Ermon \cite{kuleshov2018accurate} propose a calibration method, which is guaranteed to calibrate the regression uncertainty given sufficient data. Calibration via an \textit{(1) auxiliary model} implies training a model, e.g.~isotonic regression, on top of a network so that its predicted distribution is calibrated. Its main disadvantage is that it is not suitable for fitting heavy-tailed distributions, and is prone to over-fitting \cite{chatterjee2021regret}. Laves et al.~\cite{laves2019well,laves2021recalibration} propose \textit{(2) factor scaling}, another approach which consists of scaling the predicted uncertainty using a single scalar value $s$. The latter is optimized using gradient descent with respect to the NLL on the validation dataset. Method \textit{(2)} is more suitable for embedded applications and requires less data than \textit{(1)}, but it has less calibration potential since one value is equally applied to all the uncertainties. Phan et al.~\cite{phan2018calibrating} adapt method \textit{(1)} for the localization of single objects, and show that it results in more reliable uncertainty estimates. Part of their future work and Kraus et al.'s~\cite{kraus2019uncertainty} is to extend it to multiple-object detection; as addressed in this work.

\section{Method}
\label{sec:methds}
This section presents our approach to loss attenuation in EfficientDet \cite{tan2020efficientdet} and outlines its decoding process. Furthermore, it introduces our uncertainty propagation methods, and explains our extensions for uncertainty calibration in localization tasks. The proposed methods are model agnostic, i.e. they are identically applicable to any other object detector. 
\subsection{Uncertainty Estimation}
\label{sec:LAE}
The loss attenuation introduced by Kendall and Gal \cite{kendall2017uncertainties} is defined as follows:
\begin{equation}
\mathcal{L}_{\mathrm{NN}}=\frac{1}{2N}\sum_{i=1}^N\frac{\|\mathbf{y}_i^*-\mathbf{f}(\mathbf{x}_i)\|^2}{\sigma(\mathbf{x}_i)^2}+\log\sigma(\mathbf{x}_i)^2   
\label{eq:losssatt}
\end{equation}
with $N$ samples, ground truth $\mathbf{y}^*$, variance $\sigma(\mathbf{x})^2$ and output $\mathbf{f}(\mathbf{x})$ for input $\mathbf{x}$. 

The output of the localization head in anchor-based object detectors consists of four variables: the anchor-relative object center coordinates ($\mathbf{\hat{x}}, \mathbf{\hat{y}}$), width $\mathbf{\hat{w}}$, and height $\mathbf{\hat{h}}$. For the estimation of the aleatoric uncertainty, the four variables are modeled via a multivariate Gaussian distribution $\mathcal{N}(\mu,\,\sigma^{2})$ with a diagonal covariance approximation. Hence, we extend \cref{eq:losssatt} for object detection:
\begin{equation}
\mathcal{L}_{\mathrm{NN}}=\frac{1}{8N_{pos}}\sum_{i=1}^N\sum_{j=1}^4(\frac{\|y^*_{ij}-\hat{\mu_j}(\mathbf{x}_i)\|^2}{\hat{\sigma_j}(\mathbf{x}_i)^2}\\+\log\hat{\sigma_j}(\mathbf{x}_i)^2)\odot m_{i}
\label{eq:efflossatt}
\end{equation}
with $N_{pos}$ as the number of anchors with assigned ground truth in each batch of input images, and the mask $\mathbf{m}$ consisting of foreground ground truth boxes $\mathbf{m}=[\mathbf{y}^*\neq0]$. These features are specific for the EfficientDet baseline loss. 

\subsection{Uncertainty Propagation}
\label{sec:UP}
The default decoding process of the localization output in EfficientDet is similar to other anchor-based object detectors such as SSD \cite{liu2016ssd} and YOLO \cite{redmon2016you}. The final coordinates ($\mathbf{y}, \mathbf{x}, \mathbf{h}$ and $\mathbf{w}$) are computed via two post-processing steps. The first step consists of transforming the anchor-relative center coordinates $\mathbf{\hat{x}}, \mathbf{\hat{y}}$, width $\mathbf{\hat{w}}$ and height $\mathbf{\hat{h}}$ based on the center coordinates $\mathbf{x_a}, \mathbf{y_a}$, width $\mathbf{w_a}$, and height $\mathbf{h_a}$ of the corresponding anchor:
\begin{equation}
\begin{aligned}
  \mathbf{y} &= \mathbf{\hat{y}}\odot \mathbf{h_a} + \mathbf{y_a} \quad &  \mathbf{h} &= \exp(\mathbf{\hat{h}})\odot \mathbf{h_a} \\
  \mathbf{x} &= \mathbf{\hat{x}}\odot \mathbf{w_a} + \mathbf{x_a} \quad & \mathbf{w} &= \exp(\mathbf{\hat{w}})\odot \mathbf{w_a} 
   \end{aligned}
  \label{eq:rel_anch}
\end{equation}
\cref{eq:rel_anch} is calculated for each prediction in the five feature maps, resulting in $A_{cell}\cdot(\frac{I_H\cdot I_W}{128^2}+\frac{I_H\cdot I_W}{64^2}+\frac{I_H\cdot I_W}{32^2}+\frac{I_H\cdot I_W}{16^2}+\frac{I_H\cdot I_W}{8^2})$ iterations, with $A_{cell}$ as the number of anchors per grid cell, $I_H$ as the height of the input image and $I_W$ as its width. The decoding process yields coordinates that are relative to the scaled input image rather than the corresponding anchors. As a result, the second step consists of linearly rescaling the decoded coordinates to the original image size.

Sampling is the only approach in existing works that enables the transformation of a distribution via a non-linear function such as the exponential in \cref{eq:rel_anch}. It however either increases computation time or reduces accuracy. We therefor present two novel, exact and fast methods for decoding, via \textit{(1) normalizing flows} and via  \textit{(2) properties of the log-normal distribution}.

\textbf{(1) Decoding via Normalizing Flows.} As explained by Kobyzev, Prince and Brubaker \cite{kobyzev2020normalizing}, a normalizing flow is a transformation of a probability distribution via a sequence of invertible and differentiable mappings. The density of a sample in the transformed distribution can be evaluated by computing the original density of the inverse-transformed sample, multiplied by the absolute values of the determinants of the Jacobians for each transformation: 
\begin{equation}
p_{\mathbf{Y}}(\mathbf{y}) =p_{\mathbf{Z}}(\mathbf{f}(\mathbf{y}))|\operatorname{det} \mathbf{Df}(\mathbf{y})|=p_{\mathbf{Z}}(\mathbf{f}(\mathbf{y}))|\operatorname{det} \mathbf{Dg}(\mathbf{f}(\mathbf{y}))|^{-1}
\end{equation}

where $\mathbf{Z}\in\mathbb{R^D}$ is a random variable with a known and tractable probability density function $p_\mathbf{Z} : \mathbb{R^D}\rightarrow \mathbb{R}$, 
$\mathbf{g}$ is an invertible function, $\mathbf{f}$ is the inverse of $\mathbf{g}$, $\mathbf{Y} = \mathbf{g}(\mathbf{Z})$ is a random variable, $\mathbf{Df}(\mathbf{y})=\frac{\partial\mathbf{f}}{\partial\mathbf{y}}$ is the Jacobian of $\mathbf{f}$ and $\mathbf{Dg}(\mathbf{z}) =\frac{\partial\mathbf{g}}{\partial\mathbf{z}}$ of $\mathbf{g}$. The determinant of the Jacobian of $\mathbf{f}$ captures the scaling and stretching of the space during the transformation, which ensures that the transformed distribution has the same area as the original distribution and is a valid probability density function that integrates to one. In other words, the original density $p_{\mathbf{Z}}$ is pushed forward by the function $\mathbf{g}$, while the inverse function $\mathbf{f}$ pushes the data distribution in the opposite normalizing direction, hence the name normalizing flow. \cref{eq:rel_anch} can be reformulated into four chains of transformations on normal distributions. Let $\mathbf{g}_1(\mathbf{y}),\mathbf{g}_2(\mathbf{y})$ be invertible functions; the transformation of the distributions corresponding to the width $\mathbf{\hat{w}}$ and height $\mathbf{\hat{h}}$ is written as:
\begin{equation}
\begin{aligned}
\mathbf{g}_1(\mathbf{y}) = \exp(\mathbf{y}) &\quad \mathbf{g}_2(\mathbf{y}) = \mathbf{c}\odot\mathbf{y} \\
\mathbf{h} = \mathbf{g}_2 \circ \mathbf{g}_1(\mathbf{\hat{h}}) \text{ with } \mathbf{c}=\mathbf{h_a} &\quad \mathbf{w} = \mathbf{g}_2 \circ \mathbf{g}_1(\mathbf{\hat{w}}) \text{ with } \mathbf{c}=\mathbf{w_a}\\
\end{aligned}
\label{eq:norfl}
\end{equation}

Each of the transformations in \cref{eq:norfl} is implemented with the help of bijectors, which represent differentiable and injective functions. The final coordinates and variances in the scaled image are then directly calculated from the transformed distribution. This method can also be applied for uncertainty propagation in other anchor-based object detectors such as YOLOv3 \cite{redmon2018yolov3}, by including a sigmoid function in the chain of transformations in \cref{eq:norfl}.
\begin{figure}[ht]
  \centering
  \includegraphics[width=\linewidth]{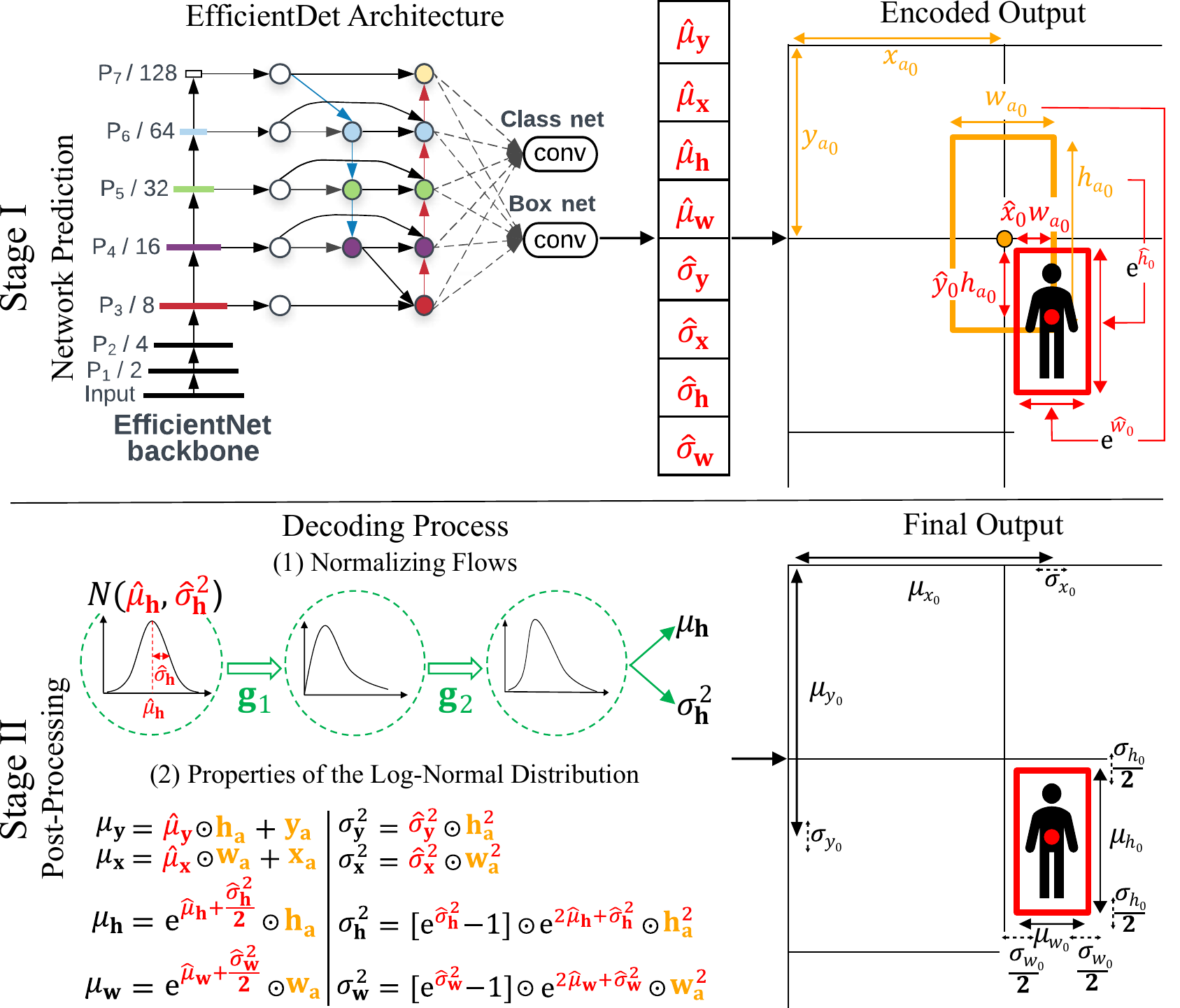}
  \caption{Illustration of our uncertainty propagation methods in the decoding process of EfficientDet\cite{tan2020efficientdet}. The output of the localization head is adjusted to predict an output distribution for each coordinate (in red). These distributions are relative to the pre-defined anchor coordinates (in orange). During post-processing, the distributions undergo non-linear transformations to obtain coordinates that are relative to the image. Both our methods provide a fast and exact propagation, with (1) allowing universality and (2) being computationally efficient. }
  \label{fig:sketch}
\end{figure}

\textbf{(2) Decoding via Properties of the Log-Normal Distribution.} The calculation of the Jacobi matrix and inverse functions is computationally expensive. We therefore introduce a different method that directly calculates the transformed mean and variance for the specific case of a normal distribution and exponential or sigmoid transformation. If $\mathbf{Z}$ follows a normal distribution with mean $\mu$ and variance $\sigma^2$, then $\mathbf{Y}=\exp(\mathbf{Z})$ follows a log-normal distribution. The density function, mean and standard deviation of a log-normal distribution are calculated as follows \cite{Balakrishnan1999}:
\begin{equation}
\begin{aligned}
f(y;\mu,\sigma^2) = &\frac{1}{y\sigma\sqrt{2\pi}}\exp(\frac{-[\log(y)-\mu]^2}{2\sigma^2})\\
Mean(\mathbf{Y}) =&\exp(\mu)\sqrt{\exp(\sigma^2)}=\exp(\mu+\frac{\sigma^2}{2})\\ SD(\mathbf{Y})=&\exp(\mu)\sqrt{\exp(\sigma^2)(\exp(\sigma^2)-1)}
\end{aligned}
  \label{eq:meanvar}
\end{equation}

Combining \cref{eq:meanvar} with \cref{eq:rel_anch} results in the transformed mean and variance for the width and height, as shown in \cref{fig:sketch}. Due to the preservation of linearity for Gaussian distributions, \cref{eq:rel_anch} remains unchanged for the mean of the center coordinates. For the variance, the equations undergo modification in compliance with the applicable transformation rules.

\textbf{Log-Normal during Training.} \cref{fig:sketch} and $Mean(\mathbf{Y})$ in \cref{eq:meanvar} show that a factor $\frac{\sigma^2}{2}$ is added to the mean of the width and height during the decoding. This always results in an enlargement of the bounding boxes ($\sigma^2>0$, $\exp{(\sigma^2)}>1)$. However, the model fits the offsets during training based solely on the mean, with no regard to the uncertainty (see \cref{eq:losssatt,eq:efflossatt}). We propose incorporating the same factor during training, thereby accounting for the exponential transformation in the decoding equations of  $\mu_\mathbf{h}$ and $\mu_\mathbf{w}$. This results in $\|y^*_{ij}-[\hat{\mu_j}(\mathbf{x}_i) + \frac{\hat{\sigma_j}(\mathbf{x}_i)^2}{2}]\|^2$ for $j=3,4$ in \cref{eq:efflossatt}.
\subsection{Uncertainty Calibration}
\label{sec:CL}
The main idea behind post-hoc calibration on the validation set is to map the uncertainty to the residuals via a model $\mathbf{r}$ or a scaling factor $s$. 

\textbf{Extensions to Caliration by a Model.} Since the model predicts a multivariate Gaussian distribution with a diagonal covariance matrix (see \cref{sec:LAE}), all four coordinates are predicted independently. Furthermore, the performance of the object detector varies from one class to the other due to heavy class imbalance, potentially leading to bias towards one class during calibration while neglecting the other. Therefore, we extend calibration via an auxiliary model \cite{vv} from calibrating all four uncertainties simultaneously with one isotonic regression model $\mathbf{r}$ to the calibration of the uncertainty for each coordinate $c$ with a separate model $\mathbf{r}_c$ for $c\in[1,4]$, each ground truth class $k$ with $\mathbf{r}_k$ for $k\in[1,n_{classes}]$, and each coordinate $i$ plus each ground truth class $k$ with $\mathbf{r}_{c,k}$. 
For an input $\mathbf{x}$, a ground truth $\mathbf{y}$ and predicted output $\mathbf{p}= \mathbf{r}(\mathbf{x})$, an isotonic regression model minimizes $\sum_{i=0}^N w_i(y_i-p_i)^2$ on $N$ predictions \cite{vv}, with $\mathbf{w}\geq0$ as the observation weight and $p_i\leq p_j$ for all $i,j \in \mathbb{E}$, where $\mathbb{E}=\{(i,j): x_i\leq x_j\}$.

\textbf{Extensions to Calibration by a Factor. } Laves et al. \cite{laves2021recalibration} optimize the factor $s$ by minimizing the NLL with gradient descent. However, the log-likelihood objective is highly sensitive towards outliers and mislabeled variables, which is particularly relevant for real-world datasets \cite{krahenbuhl2013parameter,futami2018variational}. Since their method only adjusts the predicted uncertainty $\sigma$ in $\mathcal{N}(\mu,(s\cdot\sigma)^2)$, we propose to directly optimize the scaling factor $s$ based on a distance metric between the predicted uncertainty and the true intervals, similar to the isotonic regression optimization goal. Therefore, two different loss functions are introduced, the root-mean-square uncertainty error (RMSUE) and the mean absolute uncertainty error (MAUE):
\begin{equation}
\text{RMSUE}(s)=\sqrt{\frac{1}{N}\sum_{i=1}^N\left(\Delta_i-s\cdot\sigma_i\right)^2} \quad \text{MAUE}(s)=\frac{1}{N}\sum_{i=1}^N\left|\Delta_i-s\cdot\sigma_i\right|
  \label{eq:tslosses}
\end{equation}
with $N$ detections, $\mathbf{\sigma}$ as the predicted uncertainty, and $\mathbf{\Delta} = |\mathbf{y}^*-\mu|$ as the residual. 

\textbf{Relative Uncertainty.} Existing methods are not attuned for localization tasks as they do not account for varying aspect ratios and sizes of bounding boxes. We introduce relative calibration, which consists of calibrating $\sigma$ and $\Delta$ after normalization with the width and height of their corresponding object. This prevents the uncertainty of large objects from negatively influencing the calibration of the uncertainty of smaller objects. Contextualizing the uncertainty with respect to its object also helps mitigate the effect of missing depth information in 2D images, which is crucial for the comprehension of a detector's confidence in real-world detections.

\textbf{Proximity-based Data Sorting.} Post-hoc calibration is performed on the validation set. The output of non-maximum suppression (NMS) in object detectors typically involves selecting top $n$ detections based on their classification score using a manually specified threshold, resulting in the exclusion of certain detections. Such exclusions, in turn, could correspond to actual ground truths and therefore can impede the calibration of the localization uncertainty. EfficientDet employs soft-NMS, which entails the adjustment and subsequent sorting of its output based on the classification score. Nevertheless, a higher score does not necessarily imply a more accurate detection. We propose resorting the NMS output based on the nearest-neighbor to the ground truth via a distance metric, such as mean squared error (MSE), hence retaining and correctly allocating all samples in the validation set.
\section{Experiments}
\label{sec:exper}
 The datasets used in this work are common in autonomous driving research: KITTI \cite{Geiger2012CVPR} (all 7 classes, $20\%$ split for validation), and BDD100K \cite{yu2020bdd100k} (all 10 classes, 12.5\% official split). The baseline is EfficientDet-D0 \cite{tan2020efficientdet} pre-trained on COCO \cite{cocodataset} and fine-tuned on the two datasets respectively for 500 epochs with 8 batches and an input image resolution of 1024x512 pixels. The default hyperparameters for EfficientDet-D0 are maintained. To prevent the classification results from affecting the localization output, we use ground truth classes for the per-class calibration and reorder the detections based on MSE (the distance measure used during training, see \cref{eq:efflossatt}) for both calibration and evaluation.
 
\subsection{Decoding Methods}
To showcase the effectiveness of the presented methods, eight metrics are selected. For localization: Average Precision (AP), root-mean-square error (RMSE), mean intersection over union (mIoU) and average time: model exporting time (ET) in seconds (s) and inference time (IT) in milliseconds (ms) per image. For uncertainty: RMSUE, expected calibration error (ECE) \cite{kuleshov2018accurate}, negative log-likelihood (NLL) and sharpness (Sharp). Sharpness is the average of the variance, i.e.~it relates to the concentration of the predictive distribution \cite{gneiting2007strictly}. Each model is trained three times. The results of sampling and IT are averaged over three trials on the validation set. ET is calculated as the average of three exporting iterations. Time measurements are performed on one GPU (RTX 3090). We compare our normalizing flows (N-FLOW) and log-normal (L-NORM) approaches to the baseline without uncertainty, and to the sampling method (SAMP) with 30, 100 and 1000 samples, inspired by Le et al.~\cite{le2018uncertainty}. We also add false decoding (FALSEDEC), where both $\mu$ and $\sigma$ are decoded via \cref{eq:rel_anch}, as an ablation study to analyze the effect of correct propagation and including the uncertainty in the decoding process of the mean (see \cref{eq:meanvar}). The N-FLOW method is implemented using the library TensorFlow Probability \cite{dillon2017tensorflow}.
\begin{table}[ht]
  \caption{KITTI (top) and BDD100K (bottom): Comparison between EfficientDet-D0 baseline and model with uncertainty. Our propagation methods are faster and more accurate than sampling. Uncertainty estimation increases localization performance and reduces computation time.}
  \label{tab:decodkitti}
  \centering
\resizebox{0.9\textwidth}{!}{%
  \begin{tabular}{@{}lcccccccc@{}}
    \toprule
    \textbf{Method} & \textbf{AP$\uparrow$ }& \textbf{RMSE$\downarrow$ }& \textbf{mIoU$\uparrow$ }& \textbf{NLL$\downarrow$ }& \textbf{ET$\downarrow$}& \textbf{IT$\downarrow$}\\
    &&&&& \textbf{(s)}& \textbf{(ms)}\\
    \midrule
    Baseline & 72.8 $\pm$ 0.1 &  \textbf{5.07 $\pm$ 0.1} & 90.1 $\pm$ 0.1 & - & \textbf{115.6 $\pm$ 3} &  34.8 $\pm$ 4 \\[0.1cm]
    FalseDec & 73.1 $\pm$ 0.5 & 5.27 $\pm$ 0.1 & 90.3 $\pm$ 0.1 & 4.27 $\pm$ 0.1 & 116.0 $\pm$ 3 & 31.1 $\pm$ 3 \\[0.1cm]
    L-norm & \textbf{73.3} $\pm$ 0.5 & 5.17 $\pm$ 0.2  & 90.3 $\pm$ 0.0 & 3.22 $\pm$ 0.0 & \textbf{115.6 $\pm$ 2} & \textbf{31.0 $\pm$ 3} \\
    N-flow & \textbf{73.3} $\pm$ 0.5 & 5.17 $\pm$ 0.2  & 90.3 $\pm$ 0.0 & 3.22 $\pm$ 0.0 & 116.6 $\pm$ 1 & 31.6 $\pm$ 3\\[0.1cm]
    Samp30 & 68.6 $\pm$ 0.4 & 5.43 $\pm$ 0.1 & 88.7 $\pm$ 0.1 & \textbf{3.19 $\pm$ 0.0} & 118.8 $\pm$ 2 & 34.5 $\pm$ 3 \\
    Samp100 & 71.8 $\pm$ 0.5 & 5.23 $\pm$ 0.1 & 90.1 $\pm$ 0.0  & 3.20 $\pm$ 0.0 & 117.4 $\pm$ 4 & 47.0 $\pm$ 3\\
    Samp1000 & 73.1 $\pm$ 0.5 & 5.18 $\pm$ 0.2 & \textbf{90.4 $\pm$ 0.0} & 3.21 $\pm$ 0.0 & 117.9 $\pm$ 4 & 187.4 $\pm$ 4\\ [0.1cm]

    \midrule

    Baseline &\textbf{ 24.7 $\pm$ 0.1} & 8.96 $\pm$ 0.2 & 66.6 $\pm$ 1.6 & - & 115.7 $\pm$ 3 &  33.0 $\pm$ 4  \\[0.1cm]
    FalseDec & 23.9 $\pm$ 0.2 & 8.81 $\pm$ 0.2 &  67.3 $\pm$ 0.0  & 4.40 $\pm$ 0.1   & 115.9 $\pm$ 2 & \textbf{30.4 $\pm$ 4}  \\[0.1cm]
    L-norm & 24.4 $\pm$ 0.1 & \textbf{8.53 $\pm$ 0.2} & \textbf{67.7 $\pm$ 0.0} & \textbf{3.69 $\pm$ 0.0}  & \textbf{115.3 $\pm$ 1} & 30.6 $\pm$ 4 \\
    N-flow & 24.4 $\pm$ 0.1 & \textbf{8.53 $\pm$ 0.2} & \textbf{67.7 $\pm$ 0.0} & \textbf{3.69 $\pm$ 0.0} &  116.4 $\pm$ 1 & 31.0 $\pm$ 3 \\[0.1cm]
    Samp30 & 21.0 $\pm$ 0.1  & 9.02 $\pm$ 0.2 & 64.7 $\pm$ 0.0 & 3.70 $\pm$ 0.0   & 118.0 $\pm$ 3 & 33.6 $\pm$ 4 \\
    Samp100 & 23.2 $\pm$ 0.1 & 8.68 $\pm$ 0.2 & 66.7 $\pm$ 0.0 & \textbf{3.69 $\pm$ 0.0} & 117.0 $\pm$ 3 & 45.4 $\pm$ 4 \\
    Samp1000 & 24.2 $\pm$ 0.1 & 8.55 $\pm$ 0.2 &  67.6 $\pm$ 0.1 & \textbf{3.69 $\pm$ 0.0}  & 118.4 $\pm$ 3 & 187.3 $\pm$ 4 \\
     \bottomrule
  \end{tabular}
  }
\end{table}

\textbf{Baseline vs Uncertainty.} Predicting the localization aleatoric uncertainty increases the original 3,876,321 parameters by only 2,327 (0.06\%). It reduces the required inference time per image, due to the Tensor Cores in the GPU utilizing the extension of the model output to eight values (mean and variance) \cite{mpi-forum}. The exporting time varies by decoding function. Direct calculation functions (Baseline, FALSEDEC, L-NORM) are faster than distribution-based (N-FLOW, SAMPL) functions, due to lower complexity of operations in the graph. Estimating the uncertainty improves the baseline AP and mIoU by $0.5\%$ on KITTI. On BDD100K, it reduces the AP by $0.3\%$, but improves both the mIoU and RMSE, as seen in \cref{tab:decodkitti}. Therefore, on both datasets, the localization performance increases. The COCO-style AP is affected by the classification performance, since it is calculated per class and detections are sorted based on their classification score to determine the cumulative true and false positives. This is amplified in the case of BDD100K, due to the larger number of images and their lower fidelity, and by extension, the overall decrease in performance and higher misclassification rate (see \cref{fig:kittiboxiou}) in comparison to KITTI. 

\textbf{Our Methods vs Sampling.} The only difference between the N-FLOW and the L-NORM approaches is the processing time, due to different mathematical complexity (see \cref{sec:UP}). The main advantage of the N-FLOW approach is the flexibility in changing the distribution or the transformations without manually recalculating the posterior distribution. The latter is especially beneficial, when the transformations render the posterior distribution intractable. \cref{tab:decodkitti} shows that incorrectly propagating the mean and variance (FALSEDEC) reduces performance and the precision of the uncertainty. Compared to our methods, sampling shows on both datasets either a strong reduction in performance (up to 3\% AP and mIoU) or a longer inference time per image (up to 5 times slower). However, sampling with 30 samples does offer slightly sharper uncertainties on KITTI, which results in a lower NLL. The opposite is true for BDD100K. This can be retraced to the overestimation of the uncertainty by the model. Therefore, any reduction in the uncertainty leads to an enhancement of its precision. Sampling with a mere 30 samples can result in substantial deviation in both directions, hence the fluctuation between the datasets. Based on the results in \cref{tab:decodkitti}, we select the L-NORM decoding method for the calibration evaluation.
 
\subsection{Calibration Evaluation}
Calibration improves reliability and interpretability of predicted uncertainties by reducing misalignment between the error distribution and the standard Gaussian distribution.
This is highly relevant for safety-critical applications, where uncertainty should reflect the true outcome likelihood.

\textbf{Uncertainty Behavior.} We notice that EfficientDet predicts a lower $\sigma$ on the validation set, despite the higher NLL and RMSE compared to the training set, in accordance with Laves et al.~\cite{laves2021recalibration}. We also found that $\sigma^2$ is predicted higher than the MSE, hence being miscalibrated. Reasons therefor can be found in the optimization of multiple losses and in uneven data distribution. For both datasets, the model overestimates the uncertainty, with the interval $\mu\pm\sigma$ containing $99\%$ of the true values instead of the expected $68.27\%$. 

\textbf{Calibration Methods.} For factor scaling (FS), gradient descent is applied for 100 optimization epochs with a learning rate of 0.1 on the validation dataset. Optimizing the factor $s$ based on MAUE and RMSUE (see \cref{sec:CL} and \cref{eq:tslosses}) results in a lower ECE and sharper uncertainties, but a higher NLL (see \cref{tab:kittits}). We discover a trade-off between the ECE and NLL, since optimizing $s$ based on the NLL instead results in a higher ECE. For the auxiliary isotonic regression (IR) model, we compare its extensions to per-coordinate (PCo) and per-class (CL) calibration. An illustrative example is featured in \cref{fig:godfig}. \cref{tab:kittits} shows that per-coordinate calibration outperforms the calibration on all coordinates as expected, since all four normal distributions are assumed to be independent. Per-class calibration further reduces the ECE, RMSUE and NLL, since both datasets contain heavily unbalanced classes with different aspect ratios and localization accuracy. IR outperforms FS for both datasets, because the size of the calibration dataset is large enough for the auxiliary model to train on, as also observed by Feng et al.~\cite{feng2019can}. Relative calibration results in further improvement for IR in both NLL and ECE. Our hypothesis in \cref{sec:CL} is that relative calibration mitigates bias towards larger objects. We empirically demonstrate that it effectively achieves this objective by conducting a comparative analysis on small, medium and large objects based on their area as defined by the COCO API \cite{cocodataset}. Our findings reveal that relative calibration causes a more substantial reduction in ECE on small objects with a 6-fold further decrease when compared to absolute calibration, whereas it is 2-fold on medium objects and 3-fold on large objects. Accordingly, relative isotonic regression per-coordinate and per-class (Rel. IR PCo CL) is selected for further investigations.
\begin{table}[ht]
  \caption{KITTI (left) and BDD100K (right): Comparison between different calibration methods. Our factor scaling (FS) losses outperform NLL. Isotonic regression (IR) per-coordinate (PCo) and per-class (CL) outperforms classic one-model IR and all FS approaches. Relativity increases calibration performance.}
  \label{tab:kittits}
  \centering
\resizebox{\textwidth}{!}{%
  \begin{tabular}{@{}l|ccccl@{}}
    \toprule
    \textbf{Method} & \textbf{ RMSUE$\downarrow$} & \textbf{ECE$\downarrow$}& \textbf{NLL$\downarrow$}& \textbf{Sharp$\downarrow$ } \\
    \midrule
    Uncalibrated &  13.0 $\pm$ 0.0 & 0.384 $\pm$ 0.000 & 3.22 $\pm$ 0.0 & 14.9 $\pm$ 0.0 \hspace{0.1cm}\\[0.1cm]
    FS MAUE &  4.6 $\pm$ 0.2 & 0.047 $\pm$ 0.001 & 3.14 $\pm$ 0.4 & \textbf{2.5 $\pm$ 0.0} \hspace{0.1cm}\\
    FS RMSUE & 4.6 $\pm$ 0.2 & 0.088 $\pm$ 0.003 & 2.79 $\pm$ 0.2 & 3.0 $\pm$ 0.0 \hspace{0.1cm}\\
    FS NLL & 5.0 $\pm$ 0.3 & 0.194 $\pm$ 0.021 & \textbf{2.51 $\pm$ 0.1}  & 4.7 $\pm$ 0.5 \hspace{0.1cm}\\
    Rel. FS RMSUE \hspace{0.1cm}& 7.2 $\pm$ 0.1 & 0.306 $\pm$ 0.002 & 2.74 $\pm$ 0.0 & 8.3 $\pm$ 0.1\hspace{0.1cm} \\[0.1cm]
    Abs. IR & 4.5 $\pm$ 0.2 & 0.032 $\pm$ 0.001 & 3.15 $\pm$ 0.3 & \textbf{2.5 $\pm$ 0.0}  \hspace{0.1cm}\\
    Abs. IR CL & 4.4 $\pm$ 0.2 & 0.029 $\pm$ 0.001 & 2.86 $\pm$ 0.2 & 2.7 $\pm$ 0.0 \hspace{0.1cm} \\
    Abs. IR PCo & 4.5 $\pm$ 0.2 & 0.032 $\pm$ 0.001 & 3.03 $\pm$ 0.2 & 2.6 $\pm$ 0.0\hspace{0.1cm} \\
    Abs. IR PCo CL \hspace{0.1cm}& \textbf{4.3 $\pm$ 0.2} & 0.028 $\pm$ 0.000 & 2.70 $\pm$ 0.1 & 2.9 $\pm$ 0.0  \hspace{0.1cm}\\[0.1cm]
    Rel. IR  & 4.5 $\pm$ 0.2 & 0.027 $\pm$ 0.001 & 3.06 $\pm$ 0.3 & \textbf{2.5 $\pm$ 0.0}\hspace{0.1cm} \\
    Rel. IR CL & 4.4 $\pm$ 0.2 & 0.026 $\pm$ 0.001 & 2.78 $\pm$ 0.2  & 3.1 $\pm$ 0.4\hspace{0.1cm} \\
    Rel. IR PCo & 4.5 $\pm$ 0.2 & 0.027 $\pm$ 0.001 & 3.03 $\pm$ 0.2 & \textbf{2.5 $\pm$ 0.1} \hspace{0.1cm}\\ 
    Rel. IR PCo CL \hspace{0.1cm}& 4.4 $\pm$ 0.3 & \textbf{0.025 $\pm$ 0.000} & 2.69 $\pm$ 0.2 & 3.2 $\pm$ 0.5 \hspace{0.1cm}\\
    \bottomrule
  \end{tabular}%
  \begin{tabular}{@{}|cccc@{}}
    \toprule
    \textbf{ RMSUE$\downarrow$} & \textbf{ECE$\downarrow$}& \textbf{NLL$\downarrow$}& \textbf{Sharp$\downarrow$}\\
    \midrule
     15.1 $\pm$ 0.1 & 0.323 $\pm$ 0.000 & 3.69 $\pm$ 0.0 & 17.22 $\pm$ 0.0 \\[0.1cm]
    7.5 $\pm$ 0.3 & 0.026 $\pm$ 0.001 & 4.72 $\pm$ 0.2 & 4.28 $\pm$ 0.0 \\
    7.6 $\pm$ 0.3 & 0.074 $\pm$ 0.000 & 6.43 $\pm$ 0.3 & \textbf{3.21 $\pm$ 0.0} \\
    9.4 $\pm$ 0.4 & 0.217 $\pm$ 0.008 & \textbf{3.46 $\pm$ 0.0} & 9.72 $\pm$ 0.4 \\ 
    8.5 $\pm$ 0.3 & 0.175 $\pm$ 0.003 & 3.50 $\pm$ 0.1 & 8.06 $\pm$ 0.1 \\[0.1cm]
    7.5 $\pm$ 0.3 & 0.027 $\pm$ 0.001 & 4.60 $\pm$ 0.1 & 4.09 $\pm$ 0.0 \\
    7.4 $\pm$ 0.3 & 0.026 $\pm$ 0.001 & 4.39 $\pm$ 0.1 & 4.23 $\pm$ 0.0 \\
    7.5 $\pm$ 0.3  & 0.027 $\pm$ 0.001 & 4.57 $\pm$ 0.2 & 4.11 $\pm$ 0.0 \\
    7.4 $\pm$ 0.3 & 0.025 $\pm$ 0.001 & 4.36 $\pm$ 0.1 & 4.33 $\pm$ 0.0 \\[0.1cm]
    7.4 $\pm$ 0.3 & 0.018 $\pm$ 0.001 & 4.52 $\pm$ 0.1 & 4.07 $\pm$ 0.0 \\
    \textbf{7.3 $\pm$ 0.3} & \textbf{0.017 $\pm$ 0.000} & 4.29 $\pm$ 0.1 & 4.24 $\pm$ 0.0 \\
    7.4 $\pm$ 0.3  & 0.018 $\pm$ 0.000 & 4.49 $\pm$ 0.1 & 4.08 $\pm$ 0.0 \\
    \textbf{7.3 $\pm$ 0.3} & \textbf{0.017 $\pm$ 0.000} & 4.23 $\pm$ 0.1 & 4.27 $\pm$ 0.0 \\
    \bottomrule
  \end{tabular}%
  }
\end{table}

\begin{figure}
  \centering
  \centerline{\includegraphics[width=\textwidth]{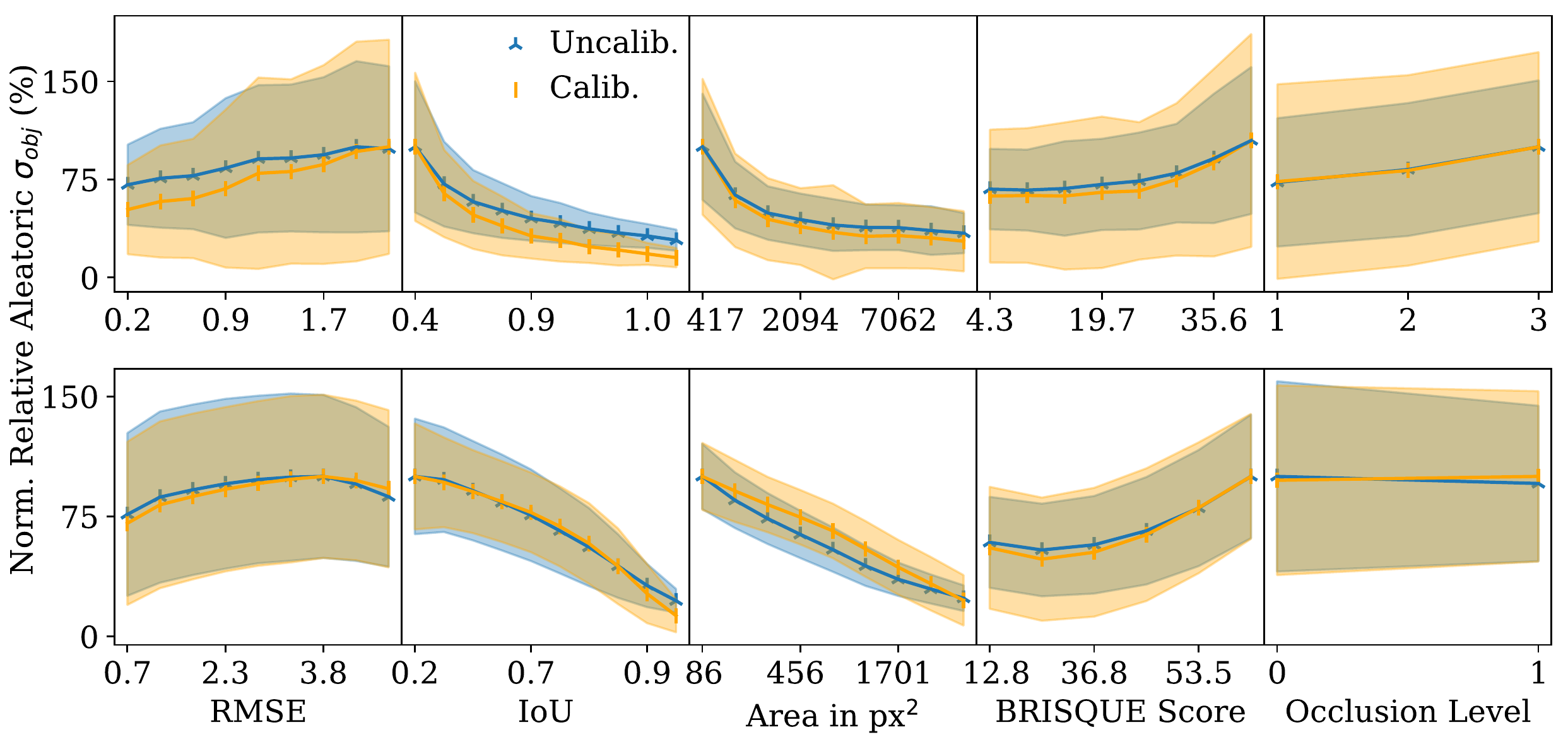}}
    \caption{KITTI (top) and BDD100K (bottom): Correlation between aleatoric uncertainty ($\mu_{\sigma_{obj}} \pm \sigma_{\sigma_{obj}}$) and performance metrics, object area, BRISQSUE score and occlusion level. Uncalibrated and calibrated via Rel. IR PCo CL, binned based on quantiles and normalized with the highest $\mu_{\sigma_{obj}}$ post-binning.}
    \label{fig:kittiarea}
\end{figure}

\subsection{Uncertainty Correlation}
We investigate the correlation between the localization aleatoric uncertainty and performance, object area, i.e.~distance in the real world, occlusion level and the Blind/Referenceless Image Spatial Quality Evaluator (BRISQUE) \cite{mittal2011blind}. In the following, $\sigma_{obj}=\frac{1}{4}\sum^4_{i=1}\sigma_i$ is the average of all four uncertainties per object.

\textbf{Uncertainty vs Real-World Metrics.} We assume that the distance of an object in the real world is connected to its area in an image in pixels$^2$ (px$^2$). For both datasets, \cref{fig:kittiarea} demonstrates that the smaller the object in the image, or the farther away it is, the higher its aleatoric uncertainty. As mentioned in \cref{sec:relw}, aleatoric uncertainty correlates with occlusion. \cref{fig:kittiarea} visualizes the results based on the annotations for occlusion in both datasets. KITTI has three occlusion levels: 0 is fully visible, 1 is occluded less than $50\%$ and 2 is occluded more than $50\%$. BDD100K has only two: 0 is visible and 1 is occluded. The correlation is present in both datasets, but less for BDD100K. We trace this back to the model predicting double the uncertainty on average for the traffic light and sign classes, as compared to other classes. While 56568 instances of these classes were labeled as visible, only 5040 were labeled as occluded (8\%). This, combined with the high uncertainty, negatively impacts the correlation. However, when excluding these two classes, the average uncertainty of visible objects is 34\% lower than occluded objects pre-calibration and 40\% lower post-calibration.

\textbf{Uncertainty vs Image Quality.} The assumption that aleatoric uncertainty correlates with inherent noise in the data is investigated based on the BRISQUE score. For every detection, the score is calculated on the standardized crop around its bounding box in the corresponding image. Standardizing crops involves mean subtraction and division by the standard deviation of pixel values. As \cref{fig:kittiarea} shows, the BRISQUE score positively correlates with the uncertainty, indicating a higher uncertainty for lower image quality.

\begin{figure}
   \centering
  \begin{subfigure}{0.49\textwidth}
   \centering
      \includegraphics[width=1.07\textwidth]{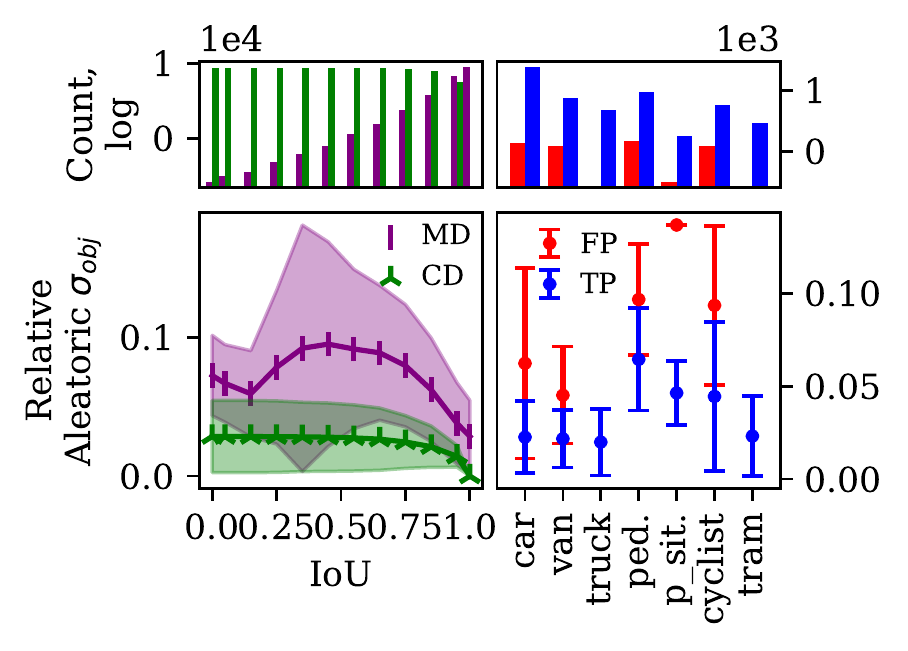}
  \end{subfigure}
  \begin{subfigure}{0.49\textwidth}
   \centering
      \includegraphics[width=0.94\textwidth]{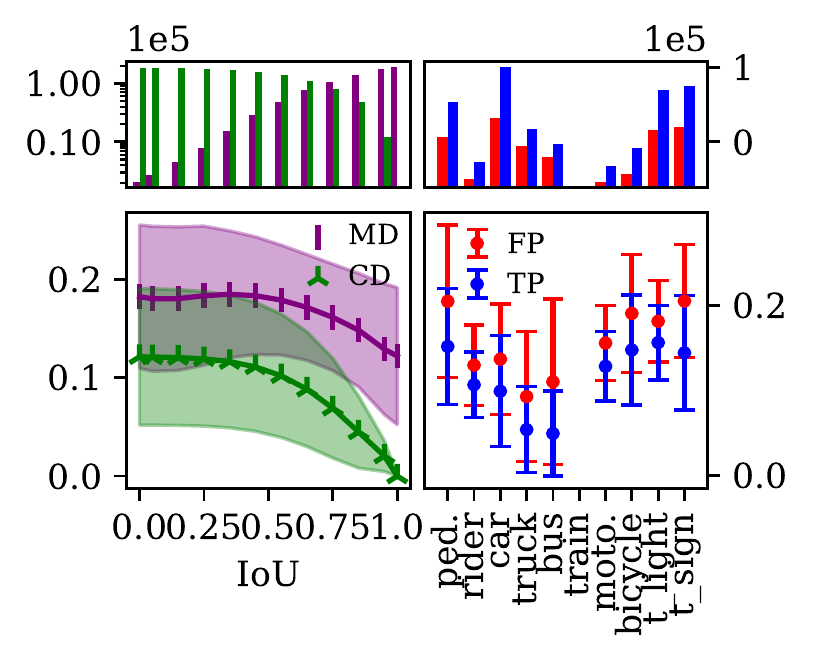}
  \end{subfigure}
    \caption{KITTI (left) and BDD100K (right): Calibrated aleatoric uncertainty ($\mu_{\sigma_{obj}} \pm \sigma_{\sigma_{obj}}$) for misdetections (MD, IoU $<=$ threshold) and correct detections (CD, IoU $>$ threshold) at each IoU threshold, and true positives (TP) and false positives (FP) for the classification of each class. The uncertainty of MDs and FPs is on average higher than CDs and TPs.}
  \label{fig:kittiboxiou}
\end{figure}

\textbf{Uncertainty vs Detection Performance.} The comparison with IoU and RMSE for both datasets in \cref{fig:kittiarea} shows a correlation with localization accuracy. Calibration via Rel. IR PCo CL strengthens the correlation with all metrics, as presented in \cref{fig:kittiarea}. The calibrated aleatoric uncertainty can be used for thresholding between misdetections (IoU $<=$ threshold) and correct detections (IoU $>$ threshold) for both datasets (see \cref{fig:kittiboxiou}), since the uncertainty of misdetections is on average higher than the uncertainty of correct detections. This extends to classification, where the uncertainty of false positives of each class is on average also higher than the uncertainty of true positives. Therefore, the localization aleatoric uncertainty allows for the detection of the model prediction errors.
\section{Conclusion}
We provide an object detection pipeline with reliable and interpretable localization uncertainty, by covering the estimation, propagation, calibration, and explanation of aleatoric uncertainty. Our methods enhance the safety and reliability of object detectors without introducing drawbacks. We propose two approaches to propagation, which allow an exact and fast propagation of distributions, along the corresponding uncertainty, through non-linear functions such as exponential, sigmoid and softmax. We demonstrate the efficacy of our techniques through their implementation in the post-processing of EfficientDet as a use-case. Our propagation methods improve the localization performance of the baseline detector on both datasets KITTI and BDD100K, and decrease the inference time. They generalize to any model with a tractable output distribution requiring its transformation via invertible and differentiable functions. They particularly alleviate the disadvantages of sampling, namely either low accuracy and reproducibility or high computation time. Furthermore, we extend regression calibration to localization, by considering the relativity of the uncertainty to its bounding box, as well as per-class and per-coordinate calibration with different optimization functions. We also investigate the data selection process for calibration and propose an approach for the allocation of predictions to their corresponding ground truth, which alleviates the disadvantages of manual thresholding. We find a correlation between aleatoric uncertainty and detection accuracy, image quality, object occlusion, and object distance in the real world. We hope the methods and results presented in this paper will encourage wider adoption of uncertainty estimation in different industrial and safety-critical applications, e.g. for safer decision making via more reliable detection monitoring, and more efficient use of labeled data in active learning.

%
%
%
%

\bibliographystyle{splncs04}
\bibliography{biblio}

\end{document}